\documentclass{article}
\pdfoutput=1

\usepackage[utf8]{inputenc} 
\usepackage[T1]{fontenc}    
\PassOptionsToPackage{hyphens}{url} 
\usepackage[hyperfootnotes=false]{hyperref}
\usepackage{booktabs}       
\usepackage{amsfonts}       
\usepackage{nicefrac}       
\usepackage{microtype}      
\usepackage[hang,flushmargin]{footmisc}
\usepackage[linesnumbered, ruled, vlined]{algorithm2e}
\usepackage{algpseudocode}
\usepackage[T2A]{fontenc}
\usepackage{CJKutf8}
\usepackage{graphicx}

\usepackage{enumerate}
\usepackage{caption}
\usepackage{subcaption}
\usepackage{comment}
\usepackage[export]{adjustbox}
\usepackage{mathtools}
\usepackage{amsthm}
\usepackage{authblk}
\usepackage{fullpage}
\usepackage{xspace}

\usepackage{enumitem}

\usepackage[sort&compress, numbers]{natbib}

\usepackage[HTML]{xcolor}
\usepackage[most]{tcolorbox}

\usepackage{Definitions}
\usepackage{dsfont}

\tcbset{
  mybox/.style={
    colback=white,
    colframe=black,
    arc=4pt,
    boxrule=0.5pt,
    left=4pt, right=4pt, top=4pt, bottom=4pt,
    width=0.9\textwidth,
    enhanced,
  }
}

\definecolor{mycommentcolor}{HTML}{4f9739}

\newlength{\inlineheight}
\title{Towards Human-AI Complementarity in Matching Tasks}

\author[1]{Adrian Arnaiz-Rodriguez}
\author[2]{Nina Corvelo Benz}
\author[2]{Suhas Thejaswi}
\author[1]{Nuria Oliver}
\author[2]{Manuel~Gomez-Rodriguez}

\affil[1]{ELLIS Alicante\\ Alicante, Spain\\
\{adrian, nuria\}@ellisalicante.org}
\affil[2]{Max Planck Institute for Software Systems\\ Kaiserslautern, Germany\\
\mbox{\{ninacobe, thejaswi, manuel\}@mpi-sws.org}}

\date{}

\begin{document}

\maketitle

\begin{abstract}
Data-driven algorithmic matching systems promise to help human decision makers make better matching decisions in a wide variety of high-stakes application domains, such as healthcare and social service provision.
%
However, existing systems are not designed to achieve human-AI complementarity:   decisions made by a human using an algorithmic matching system are not necessarily better than those made by the human or by the algorithm alone. Our work aims to address this gap.
%
To this end, we propose collaborative matching (\comatch), a data-driven algorithmic matching system that takes a collaborative approach: rather than making all the matching decisions for a matching task like existing systems, it selects only the decisions that it is the most confident in, deferring the rest to the human decision maker.
%
In the process, \comatch optimizes how many decisions it makes and how many it defers to the human decision maker to provably maximize per\-for\-mance.
%
We conduct a large-scale human subject study with $800$ participants to validate the proposed approach. The results demonstrate that the matching outcomes produced by \comatch outperform those generated by either human participants or by algorithmic matching on their own.
%
The data gathered in our human subject study and an implementation of our system are available as open source at \url{https://github.com/Networks-Learning/human-AI-complementarity-matching}.
\end{abstract}

\section{Introduction}
\label{sec:intro}
As AI systems become more integrated into high-stakes decision-making processes, understanding how they interact with human users has become a growing concern. There is increasing consensus that the ultimate goal of machine learning systems for decision support is to achieve human-AI complementarity~\citep{bansal2021most,steyvers2022bayesian,inkpen2023advancing,alur2024auditing,hemmer2024complementarity}. 
Human-AI complementarity aims for the decisions made by a human using a decision support system to be, in expectation, better than those made by either the human or the AI system alone. 

%
In recent years, there have been significant advances in developing decision support systems that achieve human-AI complementarity by design~\citep{straitouri2023improving, corvelo2024human,straitouri2024designing,de2024towards}. However, these advances have predominantly focused on classification tasks.
%
In this work, we focus on decision support systems designed to achieve human-AI complementarity in matching tasks. 

Matching tasks appear in a variety of high-stakes application domains, including matching refugees to resettlement locations~\citep{bansak2018improving, ahani2021placement,ahani2023dynamic,freund2023group,lee2024matchings}, patients to~appoint\-ments with clinicians~\citep{salah2022predict}, or blood and organ donors to recipients~\citep{mcelfresh2023matching,aziz2021optimal}. 
In these cases, a decision maker needs to distribute a limited set of resources, such as locations, appointments, or donations, among a pool of individuals, such as refugees, patients, or recipients, through matching decisions so that individuals will use the resources effectively. However, at the time of assignment, there is uncertainty as to whether a given individual will actually benefits from the resource. 

In this context, data-driven algorithmic matching systems leverage machine learning models to predict how effectively each individual under consideration would use a given resource.%
\footnote{In practice, one needs to measure whether an individual makes good use of a given resource using proxy variables which need to be chosen carefully to avoid perpetuating historical biases~\citep{bogen2018help,garr2019diversity,tambe2019artificial}.}
Then, they leverage these predictions to make matching decisions by solving a maximum weight bipartite matching problem~\citep{tanimoto1978some,lau2011iterative} where nodes represent individuals and resources, respectively, and edge weights represent the classifier'{}s confidence that an individual would effectively use a given resource. The goal is to find an assignment between individuals and resources that maximizes the sum of the edge weights.
%
While such algorithmic matching systems can optimize predicted outcomes, they still rely on human decision makers to achieve effective human-AI complementarity~\citep{ahani2021placement}.
 In practice, this means decision makers must be able to interpret and, when necessary, override the system's recommendations. However, there is growing evidence that decision makers often struggle to make these judgments appropriately~\citep{yin2019understanding,zhang2020effect,suresh2020misplaced,lai2023towards}, which can undermine the potential benefits of the system.
In this work, we propose a data-driven algorithmic matching system designed to achieve human-AI complementarity without requiring to understand when or how to override the system's recommendations.
%
%
In detail, our contributions in this work are three-fold: 
\begin{enumerate}
  \item \textbf{A collaborative algorithmic matching framework:} 
  We propose \emph{Collaborative Matching} (\comatch), a data-driven algorithmic matching framework that, unlike traditional matching frameworks, does not make all matching decisions in a matching task.\footnote{Throughout this paper, we refer to a \emph{matching task} as the process of assigning a subset of individuals to available time slots, and to a \emph{matching decision} as the assignment of an individual to a specific time slot within that task.} Instead, it matches only the individual–resource pairs that it is most certain about---by solving a maximum weight bipartite imperfect matching problem~\citep{ramshaw2012minimum}---and defers the remaining decisions to a human decision maker.
%
  Along the way, our framework utilizes \texttt{UCB1}, a well-known multi-armed bandit algorithm~\citep{slivkins2019introduction}, to optimize how many matching decisions within each matching task instance to defer to the human decision maker to provably maximize the performance.
  \item \textbf{Empirical validation through a large-scale human subject study:} 
  We conduct a large-scale human subject study with $800$~participants. They completed $6,400$ matching tasks, each of which involving the assignment of a subset of individuals to time slots. This resulted in a total of $80,000$ individual matching decisions across $40$ distinct task samples. In this setup, human participants have access to more information than the classifier simulating a realistic human-AI collaboration scenario, allowing us to evaluate the benefits of deferring matching decisions to human decision-makers.
  \item \textbf{Open data and implementation:} To facilitate future research and ensure reproducibility, we release the full dataset collected from our user study and an implementation of the proposed matching framework as open-source at \url{https://github.com/Networks-Learning/human-AI-complementarity-matching}.
\end{enumerate}

\bigskip

\xhdr{Further Related Work}
Our work builds upon related work on learning under algorithmic triage and multi-armed bandits.

The literature on algorithmic triage aims to develop classifiers that make predictions for a given fraction of the samples and leave the remaining to human experts, as instructed by a triage policy~\citep{raghu2019algorithmic,mozannar2020consistent, de2020regression,de2021classification,okati2021differentiable,charusaie2022sample,mozannar2023should}. 
Here, the triage policy determines who predicts each sample independently of each other. 
In contrast, the proposed system determines who makes each matching decision for each individual in a pool \emph{jointly} by solving a linear program.
In this context, note that learning under algorithmic triage has been also extended to reinforcement learning settings~\citep{balazadeh2022learning,straitouri2021reinforcement,fuchs2023optimizing,tsirtsis2024responsibility}.

Furthermore, our research contributes to an extensive line of work that uses multi-armed bandits in real-world applications~\citep{durand2018contextual, misra2019dynamic, mueller2019low, agrawal2019mnl, ding2019interactive, bouneffouf2020survey, straitouri2024designing}. Within this area of research, our work is most closely related to that of Straitouri et al.~\citep{straitouri2024designing}, which has used multi-armed bandits to optimize the performance of decision support systems for classification tasks based on prediction sets.

\bigskip

\noindent The rest of the paper is organized as follows. Section~\ref{sec:formulation} formalizes the matching task and introduces a framework for human-AI collaboration in matching tasks. Section~\ref{sec:algorithm} describes an algorithm to find the optimal number of decisions to defer to a human decision maker. Section~\ref{sec:experiments} presents an empirical evaluation of our framework using a large-scale human-subject study. Section~\ref{sec:discussion} discusses limitations of our approach and outlines directions for future work. Finally, Section~\ref{sec:conclusions} offers concluding remarks.

\section{A System for Human-AI Complementarity in Matching Tasks}
\label{sec:formulation}
We consider a matching task where, for each task instance, a decision maker needs to distribute a limited set $\Rcal$ of $k$ resources, each resource $r \in \Rcal$ with a capacity $c_r$, among a pool $\Ical$ of $n$ individuals using a given amount of information $\zb = (z_i)_{i \in \Ical} \in \Zcal^{n}$ about them.
Each individual $i \in \Ical$ can receive a single resource $r \in \Rcal$. The pool size $n$ and the capacities $c_r$ may change across task instances.
Further, each individual $i \in \Ical$ may ($y_i(r) = 1)$ or may not ($y_i(r) = 0)$ make an effective use of the resource $r$ they receive.
We say that effective use occurs when, upon receiving $r$, the individual $i$ benefits from it in a way that fulfills the intended purpose of the allocation.  The benefit depends on the application domain. For example, in job placement it might mean job retention or satisfaction; in healthcare, it could mean improved health outcomes, etc. 
Thus, the decision maker aims to make matching decisions that maximize the number of individuals who benefit from the resources.

\begin{figure}[t]
\centering
\includegraphics[width=\linewidth]{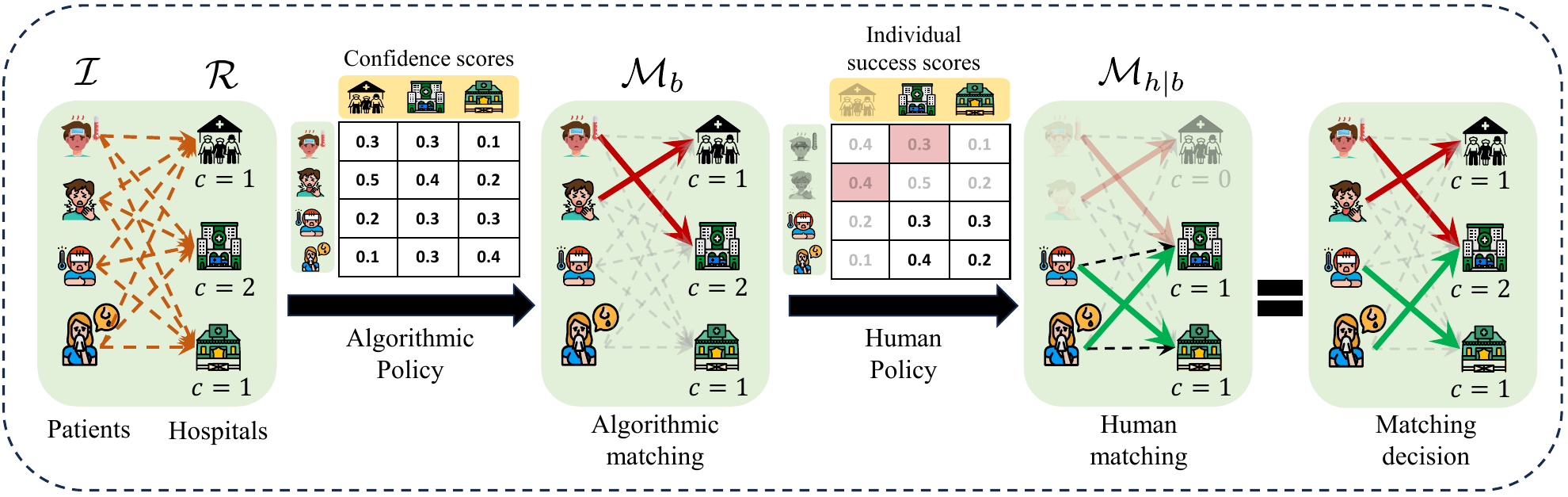}
\caption{
Our data-driven algorithmic matching framework, \comatch, is illustrated in the context of matching patients to available appointment time slots in hospitals. 
Given a pool $\Ical$ of $n$ patients and a set $\Rcal$ of time slots, the framework assists the decision maker by automatically making (feasible) matching decisions $\Mcal_b$ for $n-b$ patients, using confidence scores from a pre-trained classifier. The remaining $b$ patients are deferred to a human decision maker $h$, who completes the matching task by making the matching decisions $\Mcal_{h|b}$, using more accurate information---referred to as individual success score---on the likelihood of each patient attending an available appointment time slot.
Algorithmic assignments are indicated by red arrows and human assignments are shown in green.
}
\label{fig:system}
\end{figure}

Let $f : \Xcal \times \Rcal \rightarrow [0, 1]^{k}$ be a pre-trained classifier that, for each resource $r \in \Rcal$, 
maps an individual'{}s feature vector $x = \phi(z) \in \Xcal$, where $\phi(\cdot)$ is an arbitrary transformation%
\footnote{The transformation $\phi(\cdot)$ models the fact that, in most application domains of interest, the classifier may have access to less information about the individuals than the decision maker. Otherwise, one may argue that pursuing human-AI complementarity is not a worthy goal~\citep{alur2024auditing}. 
Accordingly, we refer to the classifier scores computed using the (restricted) features $x=\phi(z)$ as confidence scores, and to the more accurate scores computed with access to feature vector $z$ as individual success scores.}, 
to a confidence score $f_{r}(x)$, which quantifies how much the classifier believes the individual will benefit from the resource $r$.\footnote{The assumption that $f_r(x) \in [0, 1]$ is without loss of generality.}
Given the individuals' confidence scores $\fb = (f_{r}(x_i) )_{i \in \Ical, r \in \Rcal}$,
our data-driven algorithmic matching system helps the decision maker by automatically making (feasible) matching decisions $\Mcal \in 2^{\Ical \times \Rcal}$ for a subset of the individuals $\Ical' \subseteq \Ical$ using an algorithmic policy $\pi(\fb, \cb)$.
Then, the~de\-ci\-sion maker needs to make (feasible) matching decisions $\Mcal_{h} \in 2^{\Ical \backslash \Ical' \times \Rcal}$ about the individuals that have been left unmatched by the system using an unknown policy $h(\overline{\zb}, \overline{\cb})$, 
where $\overline{\cb}$ denotes the capacities left unused by the system and $\overline{\zb}$ denotes the information about the individuals left unmatched by the system.\footnote{Formally, $\overline{\zb} = (z_i)_{i \in \Ical \backslash \Ical'}$ and $(\overline{\cb})_r = c_r - |\{i \given (i, r) \in \Mcal \}|$.}
Figure~\ref{fig:system} illustrates the proposed data-driven~algo\-rith\-mic matching system \comatch.

The goal is to find the optimal algorithmic policy $\pi^{*}$ that maximizes the average number of~in\-di\-vi\-duals who benefit from the resources across pools\footnote{We denote random variables with capital letters and realizations of random variables with lower case letters.}, \ie,
\begin{equation} \label{eq:goal}
    \pi^{*} = \argmax_{\pi} \,\, \EE \left[ \sum_{(i, r) \in \Mcal} Y_i(r) + \sum_{(i, r) \in \Mcal_{h}} Y_i(r) \right].
\end{equation}
where the expectation is over the randomness in the pool generation process, the matching decisions $\Mcal$ depend on the algorithmic policy $\pi$,
and the matching decisions $\Mcal_h$ depend both on the human policy $h$ and indirectly on the algorithmic policy $\pi$.
However, to solve the above maximization problem, we need to first specify the class of algorithmic policies we aim to maximize performance upon.
Here, we consider algorithmic policies $\pi_{b}$ that make $\Mcal_b = n-b$ matching decisions by solving the following maximum weight bipartite imperfect matching problem~\citep{ramshaw2012minimum}:
\begin{equation} \label{eq:matching}
\begin{split}
    \underset{\Mcal_{b}}{\text{maximize}} \quad & \quad \sum_{(i, r) \in \Mcal_{b}} f_{r}(x_i) \\
    \text{subject to} \quad & \quad |\{r \given (i, r) \in \Mcal_{b}\}| \leq 1 \quad \forall i \in \Ical, \\
    \quad & \quad |\{i \given (i, r) \in \Mcal_{b}\}| \leq c_r \quad \forall r \in \Rcal, \\
    \quad & \quad |\Mcal_{b}| = \max(n - b, 0)
\end{split}
\end{equation}
where~$b \in \{0, \ldots, N\}$ is a parameter that controls the number of matching decisions that the system defers to the human decision maker and $N$ is the maximum pool size.

Note that, by definition, the algorithmic policies $\pi_b$ make the $n-b$ matching decisions that, 
according to the predictions made by the classifier, 
maximize the average number of individuals who would make an effective use of the resources.
Moreover, finding the optimal policy that maximizes performance reduces to the problem of finding the optimal parameter value~$b^{*}$, \ie, 
\begin{equation} \label{eq:goal-b}
    b^{*} = \argmax_{b} \,\, \EE \left[ \sum_{(i, r) \in \Mcal_b} Y_i(r) + \sum_{(i, r) \in \Mcal_{h|b}} Y_i(r) \right],
\end{equation}
where we denote the matching decisions made by the human decision maker using $\Mcal_{h|b}$ to explicitly indicate that they indirectly depend on the parameter $b$.
Importantly, the matching decisions made by a decision maker who uses the proposed system with the optimal parameter value $b^{*}$ are guaranteed to be better or equal, in expectation, than the matching decisions made by either the decision maker or the system on their own, achieving human-AI complementarity. 
This is because, under $b=N$, the human decision maker makes all matching decisions and, under $b=0$, the proposed data-driven algorithmic matching system makes all matching decisions. 

\section{Optimizing for Human-AI Complementarity}
\label{sec:algorithm}
Similarly as in the (standard) maximum weight bipartite matching problem~\citep{tanimoto1978some,lau2011iterative}, 
we can recover the solution to the imperfect matching problem defined by Eq.~\ref{eq:matching}, which may be non unique, from the solution to the following linear program:
\begin{equation} \label{eq:matching-lp}
\begin{split}
    \underset{\vb}{\text{maximize}} \quad & \quad \sum_{i \in \Ical, r \in \Rcal} f_{r}(x_i) \, v_{ir} \\
    \text{subject to} \quad & \quad \sum_{r \in \Rcal} v_{ir} \leq 1 \quad \forall i \in \Ical, \\
    \quad & \quad \sum_{i \in \Ical} v_{ir} \leq c_r \quad \forall r \in \Rcal, \\
    \
    \quad & \quad \sum_{i \in \Ical, r \in \Rcal} v_{ir} = \max(n - b, 0) \\
    \quad & \quad v_{ir} \geq 0 \quad \forall i \in \Ical, r \in \Rcal. \\
\end{split}
\end{equation}
In particular, the above linear program is guaranteed to have an optimal integral solution $\vb^{*} = (v^{*}_{ir})_{i \in \Ical, r \in \Rcal} \in \{0, 1\}^{n \cdot k}$, as shown in Appendix~\ref{app:proof-integrality}, and thus it holds that $\Mcal_{b} = \{(i,\argmax_{r \in \Rcal} v^{*}_{ir}) \given i \in \Ical \wedge \exists r \in \Rcal, v^{*}_{ir}>0\}$.
In this context, it is also worth noting that there exist specialized algorithms to find an optimal integral solution in polynomial time~\citep{ramshaw2012minimum}.

Now, since we know how to find the algorithmic policy $\pi_b(\fb, \cb)$ for any parameter value $b \in \{0, \ldots, N\}$, 
we look at the problem of finding the optimal parameter value $b^{*}$ defined in Eq.~\ref{eq:goal-b} from the perspective of multi-armed bandits from online learning~\citep{slivkins2019introduction}.
In our problem, each arm corresponds to a different parameter value $b$ and, 
at each round $t$ the system solves a matching task instance,
 where the algorithmic policy makes matching decisions~$\Mcal_{b_t}$ about $n - b_t$ individuals, 
a (potentially different) decision maker makes matching decisions $\Mcal_{h|b_t}$ about the remaining $b_t$ individuals,
and the decision maker obtains a reward $\sum_{(i, r) \in \Mcal_{b_t}} y_i(r) + \sum_{(i, r) \in \Mcal_{h|b_t}} y_i(r)$.
Then, the goal is to find a sequence of parameter values $\{ b_t \}_{t=1}^{T}$
with desirable properties in terms of total regret $R(T)$, which is given by:
\vspace{-2mm}
\begin{align}\label{eq:regret}
\begin{split}
    R(T) = T\cdot \EE &\left[ \sum_{(i, r) \in \Mcal_{b^{*}}} Y_i(r) + \sum_{(i, r) \in \Mcal_{h|b^{*}}} Y_i(r) \right] \\
    - \sum_{t=1}^{T} \EE &\left[ \sum_{(i, r) \in \Mcal_{b_t}} Y_i(r) + \sum_{(i, r) \in \Mcal_{h|b_t}} Y_i(r) \right],
\end{split}
\end{align}
where the expectation is over the randomness in the pool generation process.

To this end, we resort to UCB1 (Algorithm~\ref{alg:ucb1}), a well-known multi-armed bandit algorithm, which is guaranteed to achieve an expected regret $\EE[R(T)] \leq O(\sqrt{N t \log T})$ for any $t \leq T$, where the expectation is over the randomness in the execution of the algorithm, as shown elsewhere.
\begin{algorithm}[t]
\caption{\tt{UCB1}\label{alg:ucb1}}
\DontPrintSemicolon
\KwIn{$T, N$}

\vspace{1mm}
$t \leftarrow 0$, $\gammab \leftarrow \mathbf{0}$, $\nub \leftarrow \mathbf{0}$

\While{$t<T$} {
    \For{$b \in \{0, \ldots, N\}$} {
        $\mu(b) = \gamma(b)/\nu(b)$\\
        $\epsilon(b) = \sqrt{2 \log T / \nu(b)}$
    }
    $b_t \leftarrow \argmax_{b} \mu(b) + \epsilon(b)$\\
    $\zb_t, \cb_t \sim P(\Zb, \Cb)$ \\
    $\Mcal_{b_t} \leftarrow \pi_{b_t}(\phi(\zb_t), \cb_t)$\\
    $\Mcal_{h|b_t} \leftarrow  h(\overline{\zb}_t, \overline{\cb}_t)$\\
    $\gamma(b_t) \leftarrow \sum_{(i, r) \in \Mcal_{b_t}} y_i(r) + \sum_{(i, r) \in \Mcal_{h|b_t}} y_i(r)$\\
    $\nu(b_t) \leftarrow \nu(b) + 1$
}
\end{algorithm}

\section{Evaluation via a Human Subject Study}
\label{sec:experiments}
To empirically validate our approach, we conducted a large-scale human-subject study involving $800$ participants recruited via the Prolific platform. They completed $6,400$ matching tasks, each of which involving the assignment of a subset of individuals to time slots. This resulted in a total of $80,000$ individual matching decisions across $40$ distinct task samples.
This study was approved by the ethical review board of the faculty of mathematics and computer science at Saarland University, Germany.

In the first phase of the study, we designed a stylized setting in which participants were tasked with completing matching task instances, where they assigned patients to appointment slots in a fictional hospital, in order to evaluate how effectively human decision-makers perform under a bounded time frame. This phase established a performance baseline for human decision-making.
In the second phase, we used the data collected from this study in \comatch, which  strategically partitions the matching task between the algorithm and the human expert. By design, the resulting system aims to achieve human-AI complementarity: it produces matching outcomes whose expected utility meets or exceeds that of the human or the algorithm acting alone.

\subsection{Human Subject Study Setup}

\begin{figure}[t]
\centering
\includegraphics[width=.7\linewidth]{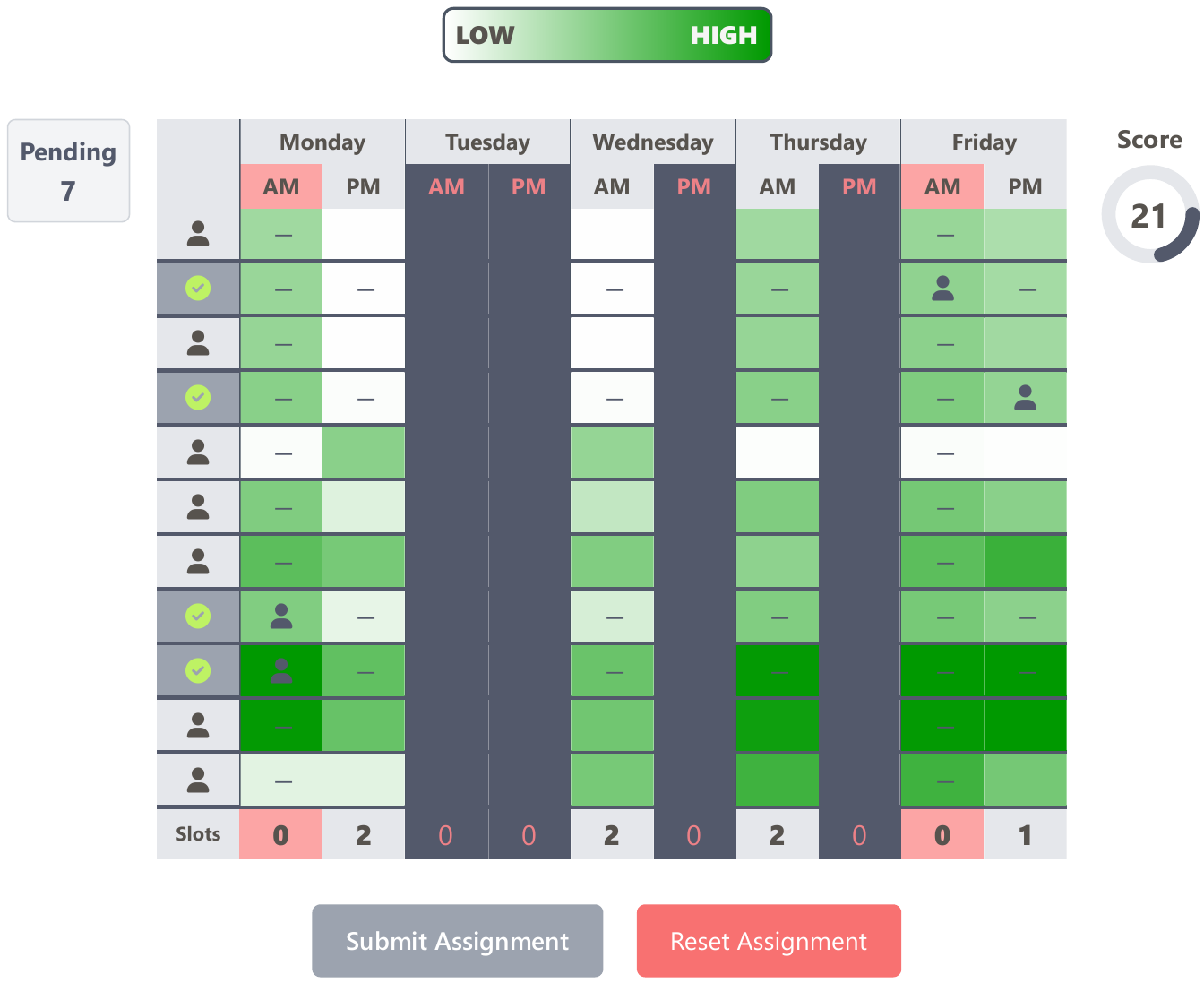}
\caption{User interface of our human subject study. The web-based user interface used on the Prolific platform is shown with an example matching task instance with $b = 11$ patients and $11$ time slots. In this example, $n-b=9$ patients were already assigned algorithmically and are not visible to the participant and the interface only displays the remaining $b=11$ matching decisions to be made by the human. In addition, a human participant has already assigned $4$ patients, indicated by the user icon placed in the time slot cells. The bottom row displays the remaining availability for each appointment slot, which is dynamically updated as participants assign patients. Individual success scores are visualized using a color gradient ranging from white to green, where white indicates low individual success score and darker shades of green indicates higher individual success score. Additional task-relevant information---such as the number of pending assignments and the current score---were shown on the left and right sides of the interface, respectively.}
\label{fig:haim-interface}
\end{figure}

The task involves matching patients on a web interface with available appointment times in a fictional hospital, with the objective of maximizing the likelihood that patients will make use of their assigned appointments. To support this, we assume access to a predictive classifier that estimates, for each patient-time-slot pair, a confidence score indicating the likelihood of attendance.

Our study was conducted using a web-based user interface and participants were recruited via the Prolific platform. The interface, shown in Figure~\ref{fig:haim-interface}, visualizes individual success scores as a heat map grid, with patients as rows and appointment slots as columns, using a color gradient ranging from white to dark green---where white indicating very low values and progressively darker shades correspond to higher individual success scores. 
The interface displays a score indicating the utility of the current matching to provide participants with an opportunity to revise their assignments to improve their score. 
To further encourage them to find optimal assignments, additional monetary incentives were offered to the top $4\%$ of the participants who consistently achieved high scores across all matching tasks.

Each matching task involves assigning a pool of $n = 20$ patients and $k = 10$ time slots, each slot with a capacity of up to $2$ patients. 
For a given matching instance, the algorithmic policy first matches $n - b$ patients by solving the maximum weight imperfect bipartite matching problem defined in Eq.~\ref{eq:matching-lp}, where the deferral parameter $b \in \{5, \dots, 20\}$ determines the number of patients that are left unmatched.
Participants were only shown the remaining $b$ patients and the matches already made by the algorithm were not visible in the interface.
By design, the classifiers' confidence scores available to the algorithm are inaccurate. Conversely, the human participant has access to more accurate individual success scores---modeling a setting in which human decision-makers have richer and more reliable information about the patients.
The remaining $b$ patients are then assigned to a human participant recruited via the Prolific platform, who should complete the matching task by making the $b$ matching decisions under a time limit of $2$ minutes per task.\footnote{The $2$-minute time limit was chosen based on pilot testing. An initial $90$-second version resulted in a high rate of incomplete submissions, especially for matching tasks involving a larger number of matching decisions. Increasing the time limit to $2$ minutes significantly improved task completion rates, offering participants a fairer opportunity to finish the assignments.}

Although submitting before the $2$-minute limit was permitted, participants could submit only after all matching decisions were completed. If the time limit was exceeded, the current assignment was automatically submitted, and the system proceeded to the next task.
Each participant completed $8$ matching task instances, with the deferral parameter $b \in \{5, \dots, 20\}$ varying across tasks.%
\footnote{In our experiments, we observed that small instances of matching problems were trivial for human participants to solve for~$b\leq 4$. Therefore, the study considered instances with larger values of $b\geq 5$.}
To ensure coverage and diversity, each participant was assigned either all even or all odd values of $b$, sampled in a uniformly random order---thereby systematically varying the number of decisions deferred to the participant across the tasks.
Further details on the generation of the matching task is provided in Appendix~\ref{app:datagen}.

The study begins with a detailed explanation of the matching task, followed by an introductory matching task that allowed participants to familiarize with the interface before starting the actual tasks. To ensure data quality, two attention checks were interleaved at the second and seventh positions to detect automated bots and verify that participants were paying adequate attention throughout the study. These attention checks were approved by experts from the Prolific platform. Failing either of these checks led to disqualification of $56$ participants out of an initial cohort of $856$, resulting in the $800$ valid participants in our study.
The participants ranged in age from $18$ to $81$ years, with a mean age of $41.9$ and a standard deviation of $13.9$. In terms of gender, $50.4\%$ identified as female and $49.6\%$ as male.

Participants who successfully completed all tasks received a base compensation of~$\pounds2$ ($\pounds8.5$ per hour). The top $4\%$ of participants, based on the highest average expected utility across all tasks, received an additional bonus of~$\pounds8$.

\subsection{Finding the Optimal Number of Decisions to Defer}

\begin{figure}[t]
\centering
   {\includegraphics[width=.60\textwidth]
    {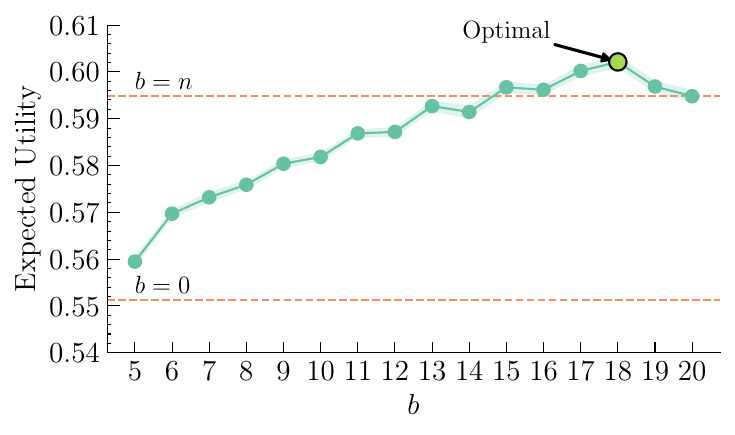}}
  \caption{Empirical expected utility per arm achieved by \texttt{UCB1} across $100$ independent realizations, each with a time horizon of $T=2000$. The two orange dotted lines, correspond to the average utility of the matching task made solely by algorithmic ($b=0$) and human ($b=n$) policy, respectively. Shaded bands denote the $95\%$ confidence intervals across the $100$ runs for each value of $b$.}
  \label{fig:bandit-all-random}
\end{figure}

Using data from the previously described stylized human-subject study, our goal is to learn the optimal number of patients $b^*$ to defer to human decision-makers so that, by design, it achieves human-AI complementarity. As presented in Section \ref{sec:algorithm}, we cast the problem of finding $b^*$ as a multi-armed bandit problem which we solve using \texttt{UCB1} algorithm.
At each time step $t$, we randomly sample a matching task instance with $n=20$ patients. The \texttt{UCB1} algorithm selects an arm $b_t$, which determines the number of matching decisions deferred to the human decision maker. The reward associated with $b_t$ is computed as the expected utility of the entire matching task, aggregating the utility of the algorithmic matching for $n-b_t$ patients and the utility of the human matching for the remaining $b_t$ patients.\footnote{When multiple human decision makers have completed the matching for a given value of $b$ for the same matching task instance, we randomly choose one of their solutions to compute the reward. If the human assignments were incomplete, the remaining unassigned patients were assigned to the available slots at random (see Appendix~\ref{app:exp-nonassign} for more details).}

Figure~\ref{fig:bandit-all-random} depicts the empirical expected utility for each value of $b$ recovered by~\texttt{UCB1}. Note that the algorithmic policy relies on inaccurate confidence scores to produce an assignment (or matching) that maximizes utility with respect to those scores. However, the resulting utility may fall short of the maximum achievable under accurate individual success scores, and can therefore be lower than the utility achieved by a human decision-maker. This is because the human decision-maker operates with different and more accurate confidence scores --- \ie, individual success scores --- reflecting a more informed understanding of the patients' conditions. This can be clearly observed in the figure the orange dotted horizontal lines depicting~$b=0$ and~$b=n$.

The expected utility increases with the value of $b$---the number of patients deferred to the human decision-maker---as human participants, equipped with the individual success scores, are able to take more informed decisions. This trend supports our hypothesis that human judgment can complement algorithmic recommendations: combining human and algorithmic matching decisions yields higher expected utility than relying solely on the algorithmic policy. The utility peaks at $b^* = 18$, beyond which the expected utility begins to decline, which suggests that deferring more than $18$ decisions to the human may surpass their cognitive capacity, leading to reduced performance.

\begin{figure}[t]
  \centering

  \begin{tabular}{c c c}
    \textbf{top-tier} & \textbf{mid-tier} & \textbf{bottom-tier} \\[2pt]

    \includegraphics[width=.32\textwidth]%
      {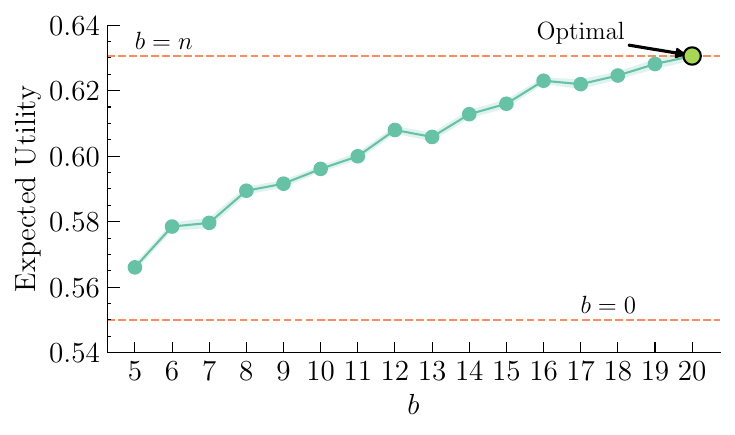} &

    \includegraphics[width=.3\textwidth,
                     trim=22 0 0 0,clip]%
      {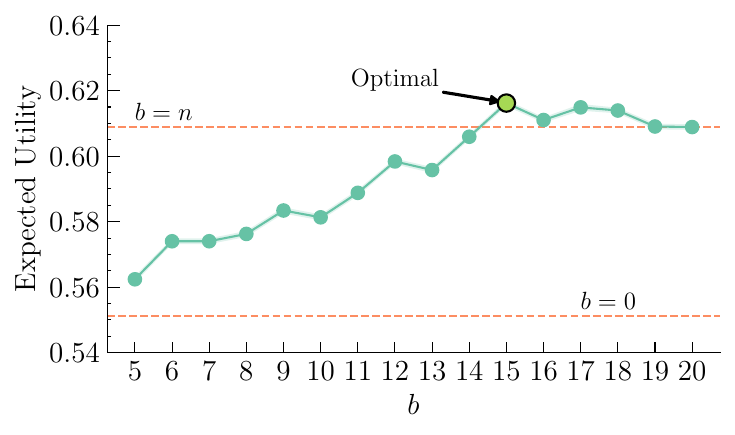} &

    \includegraphics[width=.3\textwidth,
                     trim=22 0 0 0,clip]%
      {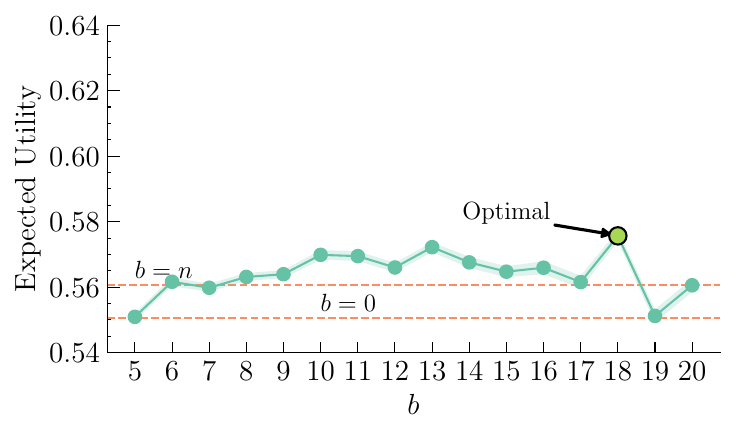}
  \end{tabular}

  \caption{Empirical expected utility per arm achieved by \texttt{UCB1} across $100$ runs (with time horizon $T=1000$) for top-, mid- and bottom-tier participants. Red dotted lines mark $b=0$ (algorithm-only) and $b=n$ (human-only); shaded bands show $95\%$ confidence intervals.}
  \label{fig:bandit-usergroups}
\end{figure}

\xhdr{Influence of Human Policy on Matching Outcomes}
A key factor that impacts the expected utility is the performance of the human decision-makers. Although the algorithm consistently returns matching decisions that maximizes the expected utility with respect to its (noisy) confidence scores, the total utility achieved by the combined human–algorithm policy depends largely on the quality of the human decisions. 
There is considerable variability across individual human decision makers, stemming from differences in cognitive ability, decision-making strategy, and subjective judgment. 
To better understand this effect, we conducted a more fine-grained analysis by categorizing human participants into three groups, top-tier, mid-tier, and bottom-tier, based on the quality of the matching decisions that they produced in the human subject study. The grouping of the participants was performed as follows. We computed the distribution of expected utility achieved by each participant, divided the distribution into deciles, and grouped participants accordingly: the bottom $40\%$ were classified as bottom-tier, the middle $20\%$ as mid-tier, and the top $40\%$ as top-tier decision-makers. This resulted in $320$, $160$, and $320$ participants in each group, respectively.

Figure~\ref{fig:bandit-usergroups} depicts the empirical expected utility for each value of $b \in \{5,\dots,20\}$, stratified by participant tier: top (left), mid (center), and bottom (right).
As expected, in the case of top-tier participants, the human policy outperforms both the algorithmic and combined policies. This improvement is due to the effectiveness of the human participants as evidenced by the high utility of the human-only policy at $b = n$, where \comatch achieves the peak utility ($b^* = 20$), suggesting that when human decision-makers are highly reliable deferring the decisions to them is beneficial. 
For mid-tier participants, the combined human–algorithm policy outperforms both standalone policies, with slightly lower average expected utility compared to top-tier participants. In this group, the peak utility is observed at $b^*=15$, indicating a more balanced strategy---sharing the decision-making load between the algorithm and the human---could be better.

Regarding bottom-tier participants, one might naturally expect that deferring more decisions to the algorithm would lead to better outcomes, given the limitations of relying on low-performing human decision-makers. However, in our study setup, the average utility achieved by human-only and algorithm-only policies in this group is relatively close, with the algorithm-only policy performing slightly worse. As a consequence, the combined human-algorithm policy achieves its peak expected utility at $b^*=18$, a slightly larger value than that obtained for the mid-tier participants.
While this may appear counterintuitive, since one might expect that as the gap between human-only and algorithm-only policies narrows, the optimal value $b^*$ should decrease, favoring fewer assignments to humans. However, we observe that the performance of participants in this tier is consistently poor across all values of $b>10$. Thus, the benefit of assigning more individuals to participants in this tier is not as pronounced. As a consequence, \ucb converges to a relatively larger value of $b^*$ compared to mid-tier participants, not because deferring more is better, but because no value of $b$ yields reliably better human performance.

\section{Discussion and Limitations}
\label{sec:discussion}
In this section, we discuss several assumptions and limitations of our work, pointing out avenues for future research. 

\xhdr{Methodology}
\comatch contributes with a promising strategy to achieve human-AI complementarity in matching tasks by learning to defer a subset of matching decisions---depending on the size of the task and the ability of the human---to human decision-makers to achieve optimal expected utility.
Our experimental results reveal that the expected utility decreases as the number of decisions are deferred to the human, suggesting that deferring based on problem size is indeed beneficial to achieve complementarity. However, task size alone does not always reflect the true complexity of the matching task. 
In many cases, the distribution of individual success scores, their proximity, and the perceptual difficulty for a human to distinguish between them may better indicate task difficulty than size alone. This highlights the need for more principled strategies that determine what to defer to humans, rather than merely how much to defer. A small matching instance may still be difficult for human decision-makers to solve, if the individual success scores are close, making them difficult to distinguish under limited time. Conversely, poorly calibrated confidence scores misrepresent the true likelihood of an individual utilizing a resource. In both cases, deferring decisions solely based on task size can be suboptimal.

\xhdr{Human Subject Study} 
During the human subject study, an important design choice was the visualization of the individual success scores: while we used a heatmap interface, alternative visual representations could influence human interpretability and decision accuracy, potentially influencing the overall utility.
Additionally, we designed the data generation process to balance realism with experimental control, aiming to create real-world scenarios while also retaining the flexibility to simulate conditions relevant to human-AI complementarity in matching tasks. 
To simulate an algorithmic matching policy that is less efficient in terms of average expected utility, we restricted the access to certain features while generating the algorithm's confidence scores, allowing us to tune---to some extent---the relative performance of human and algorithmic decision-makers. However, the setup is not without limitations. 
For example, one might expect the algorithm to outperform bottom-tier human participants, thereby creating clear conditions where deferring more decisions to the algorithm would yield better utility. This pattern did not always emerge in our experimental setting. Thus, carefully engineering the synthetic generator in future work---\eg, by varying access to features or task difficulty---could help isolate when and for whom deferral to humans or algorithms is most beneficial in matching tasks.

\section{Conclusions}
\label{sec:conclusions}

In this work, we have introduced \comatch, a hybrid decision-support framework aimed at achieving human-AI complementarity in matching tasks, with a particular focus on identifying how many decisions to defer in large-scale matching tasks. Through a large-scale human subject study, we have empirically demonstrated that deferring a subset of decisions to human decision-makers can improve expected utility---especially when those decision-makers are effective and capable. However, our findings also reveal important limitations: the utility gains from human input depends on the decision-maker’s performance, and diminishing returns emerge as cognitive load increases beyond a certain point. These insights suggest that future decision-support systems should be designed to account not only for algorithmic uncertainty, but also for the variability and cognitive limitations of human decision-makers using these systems. A promising direction for future work is the development of deferral strategies that account for which decisions to defer, rather than simply how many.

\vspace{2mm}
\xhdr{Acknowledgements} 
Gomez-Rodriguez acknowledges support from the European Research Council (ERC) under the European Union'{}s Horizon 2020 research and innovation programme (grant agreement No. 945719).
Arnaiz-Rodriguez and Oliver acknowledge support from Intel corporation, a nominal grant received at the ELLIS Unit Alicante Foundation from the Regional Government of Valencia in Spain (Convenio Singular signed with Generalitat Valenciana, Conselleria de Innovación, Industria, Comercio y Turismo, Dirección General de Innovación) and a grant by the Banc Sabadell Foundation. In addition, this work is partially funded by the European Union EU - HE ELIAS (grant agreement No. 101120237). Views and opinions expressed are however those of the author(s) only and do not necessarily reflect those of the European Union or the European Health and Digital Executive Agency (HaDEA).

{ 
\bibliographystyle{unsrtnat}
\bibliography{ai-assisted-matching}
}

\newpage
\appendix
\section{Existence of an Optimal Integral Solution to the Linear Program} \label{app:proof-integrality}

To show that the linear program defined in Eq.~\ref{eq:matching-lp} has an integral optimal solution $\vb^{*}=(v^{*}_{ir})_{i \in \Ical, r \in \Rcal} \in \{0, 1\}^{n \cdot k}$, it is sufficient to verify that the matrix $A$ associated with the constraints of the linear program is totally unimodular.

First, recall that $|\Ical|=n$ and $|\Rcal|=k$. The constraints of the linear program in Eq.~\ref{eq:matching-lp} can be written in matrix formulation as
$A\vb \leq \db, \vb \geq 0 $ for $A \in \{-1,0,+1\}^{(n+k+2)\times n\cdot k} \text{ and } \db \in \mathbb{R}^{n+k+2}$
with
\begin{equation} \label{eq:lp-matrix}
A_{i,j} = \begin{cases}
    1 & \text{if } i \leq n \text{ and } i\cdot k \leq j \leq i\cdot k + k \\
    1 & \text{if } n \leq i \leq n + k \text{ and } i \equiv j \mod k \\
    1 & \text{if } i = n+k+1 \\
    -1 & \text{if } i = n+k+2 \\
    0 & \text{otherwise}
\end{cases} 
\quad \text{and} \quad 
\db = \begin{pmatrix}
    \mathbf{1} \\ \cb  \\ \max(n-b,0) \\- \max(n-b,0) 
\end{pmatrix},
\end{equation}
where, in the vector $\vb$, we stack all the variables $v_{ir}$ associated to each individual $i \in \Ical$ in turn.
Below we give an example of matrix $A$ when $|\Ical|=3$ and $|\Rcal|=2$,
\[
\begin{pmatrix}
    1 & 1 & 0 & 0 & 0 & 0 \\
    0 & 0 & 1 & 1 & 0 & 0 \\
    0 & 0 & 0 & 0 & 1 & 1 \\
    1 & 0 & 1 & 0 & 1 & 0 \\
    0 & 1 & 0 & 1 & 0 & 1 \\
    1 & 1 & 1 & 1 & 1 & 1 \\
    -1 & -1 & -1 & -1 & -1 & -1 \\
\end{pmatrix}\,.
\]

According to Hoffman and Kruskal's theorem \citep{schrijver1998theory}, if $A$ is an integral matrix, then $A$ is totally unimodular if and only if the polyhedron $\{x \given x\geq 0; Ax\leq \db\}$ is integral for each integral vector $\db$.
As $\db$ is integral in Eq.~\ref{eq:lp-matrix}, it follows that, if $A$ is totally unimodular, then the linear program in Eq.~\ref{eq:matching-lp} has an integral optimal solution.

To show that $A$ is totally unimodular, we use the characterization of Ghoulia-Houri \citep{schrijver1998theory}. In this characterization, a matrix $A$ is totally unimodular if and only if each collection of rows of $A$ can be separated into two subsets such that the sum of the rows in one subset minus the sum in the other subset is a (row) vector with entries only in \{-1,0,1\}.\footnote{In the reference, this characterization is written for columns. However, as the transpose of a totally unimodular matrix is also totally unimodular, this characterization is also valid for rows.}

Let $A_i$ denote the $i$-th row of $A$, and let $A_\Ical$ ($A_\Rcal$) denote the subset of rows corresponding to the matching (capacity) constraints for each individual (resource), \ie, $A_\Ical = \{A_1, \dots, A_n\}$ and $A_\Rcal = \{A_{n+1}, \dots, A_{n+k}\}$. 
Let $T$ be an arbitrary subset of rows from $A$, we show that there exists a partition of $T$ into $T^+$ and $T^-$ such that 
\[
\alpha:= \left( \sum_{a \in T^+} a + \sum_{a' \in T^-} - a' \right) \in \{-1,0,1\}^{n\cdot k}.
\]
Let $\alpha^+$ denote the result of the left sum over subset $T^+$ and $\alpha^-$ denote the result of the right sum over subset $T^-$. We distinguish three cases.
\begin{itemize}
    \item If $\{A_{n+k+1}, A_{n+k+2}\} \subseteq T$ or $\{A_{n+k+1}, A_{n+k+2}\} \cap T = \emptyset$, let $T^+= T \cap (A_\Ical \cup \{A_{n+k+1}, A_{n+k+2}\})$ and $T^-= T \cap A_\Rcal$. Note that, this is a valid partition of $T$ as $A= A_\Ical \cup A_\Rcal \cup \{A_{n+k+1}, A_{n+k+2}\}$. Since rows $A_{n+k+1}$ and $A_{n+k+2}$ are either not in $T$ or they sum up to vector $\mathbf{0}^T$, we have that
    \[ \alpha^+= \sum_{a \in T^+} a = \sum_{a \in T \cap A_\Ical} a \,.\]
    Now, observe that for any $j \in [n\cdot k]$, there is only one row $a\in A_\Ical$ and only one row $a'\in A_\Rcal$ such that $a_j$, respectively $a'_j$, is non zero (\ie, $a_j=a'_j=1$).
    Thus, it follows that 
    \[ \alpha^+_j \in \{0,1\} \quad \text{and} \quad \alpha^-_j \in \{0,-1\}
    \]
    implying that $\alpha \in \{-1,0,1\}^{n\cdot k}$.
    \item If $A_{n+k+1}\in T$ and $A_{n+k+2}\not\in T$, let $T^+= T \cap (A_\Ical \cup A_\Rcal)$ and $T^-=\{A_{n+k+1}\}$.
    By the same reasoning as above, we have that $\alpha^+_j \in \{0,1,2\}$ for all $j \in [n\cdot k]$. Since $\alpha^-= -A_{n+k+1} = -\mathbf{1}^T$, it follows that $\alpha \in \{-1,0,1\}^{n\cdot k}$.
    \item If $A_{n+k+1}\notin T$ and $A_{n+k+2}\in T$, let $T^+= T \cap (A_\Ical \cup A_\Rcal \cup \{A_{n+k+2}\})$ and $T^-=\emptyset$. As $A_{n+k+2} = - A_{n+k+1}$, it follows analogously to the previous case distinction that $\alpha^+ \in \{-1,0,1\}^{n\cdot k}$. Thus, we have that $\alpha \in \{-1,0,1\}^{n\cdot k}$ as $\alpha^-=0$.
\end{itemize}
In all cases $\alpha \in \{-1,0,1\}^{n\cdot k}$, which proves that $A$ is totally unimodular, so the linear program in Eq.~\ref{eq:matching-lp} has an integral optimal solution.

\section{Data Generation Process}
\label{app:datagen}
\begin{figure}[t]
\centering
\includegraphics[width=\textwidth]{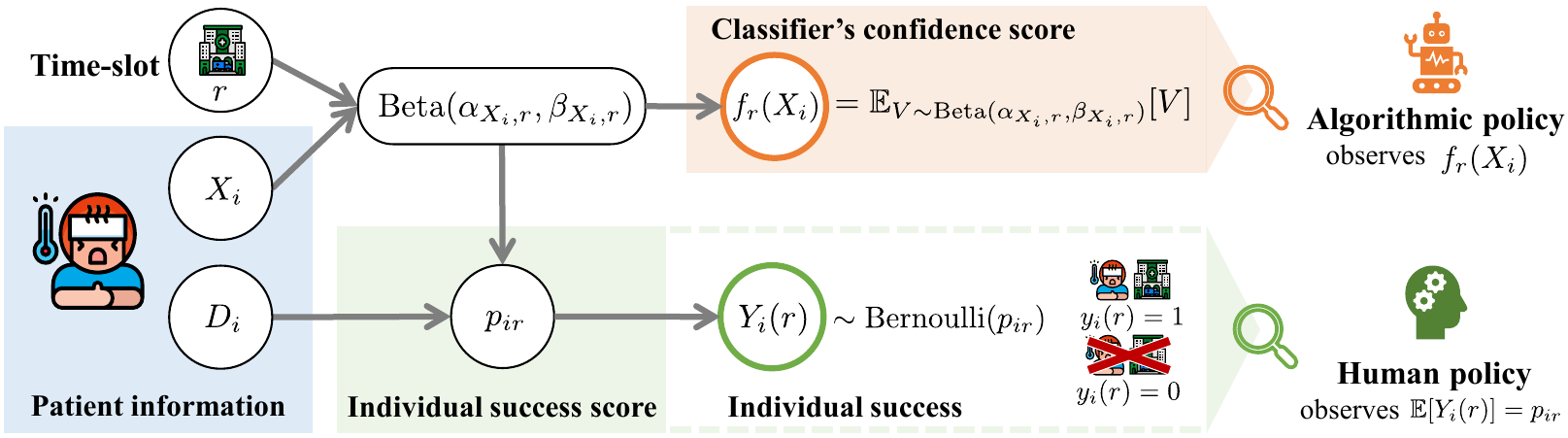}
\caption{Data generation process. Human policy observes both $X_i$ and $D_i$, and can thus use the patient's expected success $\mathbb{E}[Y_i(r)] = p_{ir} = Q_{(\alpha_{X_i,r}, \beta_{X_i,r})}(D_i)$ ---\ie, individual success score --- to perform the matching. On the other hand, the algorithmic policy does not observes~$D_i$ and uses the classifier's confidence score~$f_r(X_i) = \EE_{V \sim \mathrm{Beta}{(\alpha_{X_i,r}, \beta_{X_i,r})}}[V]$ to perform the matching.
}
\label{fig:data-gen-graph}
\end{figure}

In this subsection, we describe the synthetic data generation process used in the human subject study to generate the matching problems. We simulate a scenario in which patients must be matched to available time slots in a hospital, with each patient–slot pair associated with an individual success score indicating the likelihood that the patient will attend the assigned appointment. Each matching task consists of a pool $\Ical$ of $n=20$ patients to be assigned across a set $\Rcal$ of $k=10$ time slots, each slot $r \in \Rcal$ has a capacity of $c_{r}=2$ available appointments.%
\footnote{Each time slot $r \in \{1,\dots,10\}$ represent half-day period spanning from Monday morning to Friday afternoon \ie, \{Monday-am, Monday-pm, Tuesday-am, $\dots$, Friday-pm\}, as illustrated in Figure~\ref{fig:haim-interface}.}
An illustration of the data generation process is available in Figure~\ref{fig:data-gen-graph}.

\xhdr{Patient Information}
A widely accepted assumption in the human-AI collaboration literature is that human experts often have access to richer contextual information than what is available to machine learning models---since not all relevant information can be captured or encoded in the input features provided to the classifier. To reflect this, our data generation process is designed to produce more accurate confidence scores for human participants --- individual success scores --- while providing less accurate scores to the algorithmic matching. Each party---humans and the algorithm---then makes the assigned matching decisions using its respective scores, with the same objective of maximizing the number of patients who make use of their assigned appointments.

To model this, we represent the information of each patient $i \in \Ical$ as a pair~$Z_i=(X_i,D_i)$, where $X_i$ denotes the observable features available to both humans and the algorithm, and $D_i$ denotes additional contextual information accessible only to the human policy.%
\footnote{Recall that capital letters denote random variables, while lowercase letters represent their realizations.}
The algorithmic policy generates confidence scores based solely on $X_i=\phi(Z_i)$, where the projection $\phi$ discards the variable $D_i$ from $Z_i$, while the human policy uses the full information $(X_i, D_i)$, resulting in more accurate confidence scores. 
The variable $X_i$ is drawn from a categorical distribution $\Xcal = \{0, 1, 2\}$, with probabilities 
$P(X_i=0) = 0.20$, $P(X_i=1) = 0.45$, $P(X_i=2) = 0.35$.
The variable $D_i$ is sampled from a uniform distribution:~$D_i \sim U(0,1)$.

\begin{figure}[t]
  \centering
  \includegraphics[width=\linewidth]{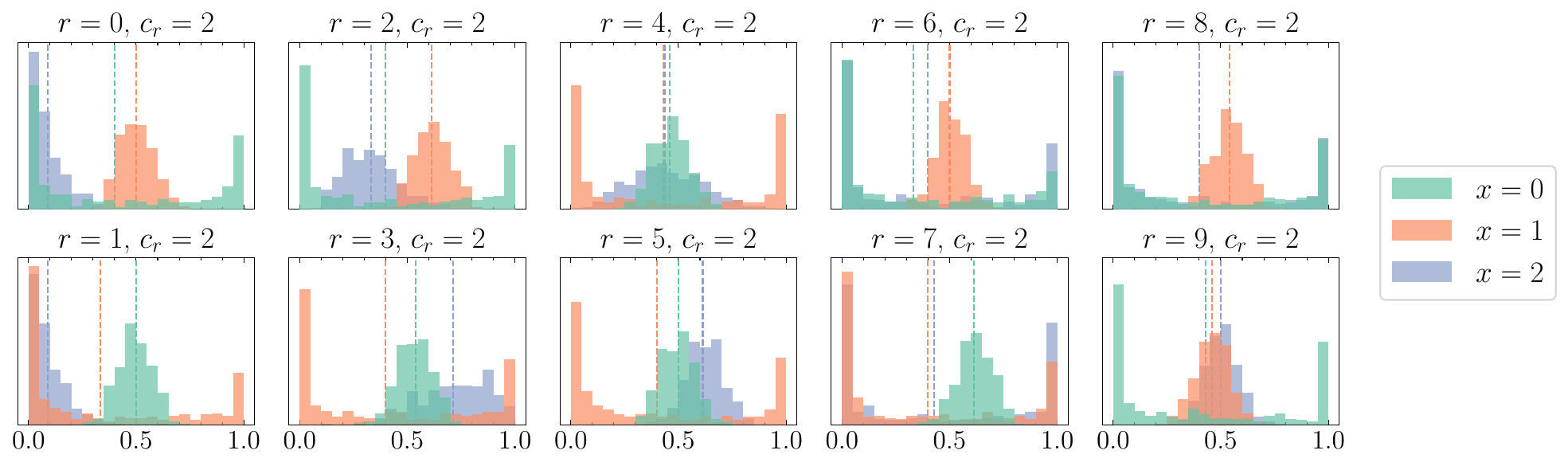}
  \caption{Beta distributions for the parameters defined in Table~\ref{tab:beta-params}. Dashed vertical lines mark the classifier's confidence scores $f_{r}(x)$ used by the algorithmic policy given $x$.}
  \label{fig:beta-params}
\end{figure}

\xhdr{Individual Success} We define individual success based on whether patient $i \in \Ical$ attends their assigned appointment at time slot $r$, \ie, $y_i(r) = 1$ if the patient attends, and $y_i(r) = 0$ otherwise.
We model $Y_i(r)$ as a Bernoulli random variable, whose probability $p_{ir}$ depends on the patient's features $Z_i=(X_i,D_i)$ and the time slot $r$.
More precisely, for every feature $x \in \Xcal$ and time slot $r \in \Rcal$, we define a Beta distribution $\mathrm{Beta}(\alpha_{x,r},\beta_{x,r})$, whose parameters~$(\alpha_{x,r},\beta_{x,r})$ depend on the specific combination of $x$ and $r$.
Table~\ref{tab:beta-params} lists the parameters of the $\mathrm{Beta}(\alpha_{x,r},\beta_{x,r})$ distribution for each $x \in \Xcal$ and $r \in \Rcal$, while Figure~\ref{fig:beta-params} illustrates the corresponding distributions.

For each patient and time slot pair $(i,r) \in \Ical \times \Rcal$, we set the variable $p_{ir}$ (i.e., the success probability for the Bernoulli distribution of $Y_i(r)$) as the $D_i$-quantile of the corresponding beta distribution:~$p_{ir}=Q_{(\alpha_{X_i, r},\beta_{X_i, r})}(D_i)$,
where $Q_{(\alpha,\beta)}$ is the quantile function (\ie, inverse cumulative distribution function) of the $\mathrm{Beta}(\alpha,\beta)$ distribution.\footnote{That is, $Q_{(\alpha,\beta)}\colon[0,1]\to[0,1]$ returns the value $p$ such that $\Pr\{V \le p\}=D_i$ for a random variable $V \sim \mathrm{Beta}(\alpha,\beta)$.}
Finally, we sample $Y_i(r)$ from a Bernoulli distribution with parameter $p_{ir}$. In sum:
\begin{equation}
 p_{ir}=Q_{(\alpha_{X_i,r},\beta_{X_i,r})}(D_i),\qquad
 Y_i(r)\sim\mathrm{Bernoulli}(p_{ir}),
\end{equation}
where the expected value for success is $\EE[Y_i(r)]=p_{ir}$.

In this study, the human participants can access $Z_i=(X_i, D_i)$. Therefore, they observe the expected individual success score, $\EE[Y_i(r)]=p_{ir}$, which is an example of an accurate confidence score.

\begin{table}
  \centering
  \caption{Beta parameters $(\alpha_{xr},\beta_{xr})$. Rows correspond to feature $x\in\{0,1,2\}$ and columns to time slots~$r\in\{1,\dots,10\}$.}
  \label{tab:beta-params}
  \resizebox{\textwidth}{!}{%
  \begin{tabular}{ccccccccccc}
    \toprule
    & \multicolumn{10}{c}{$r \in \{1,\dots,10\}$, where each $r$ represents a time slot.}\\
    & 1 & 2 & 3 & 4 & 5 & 6 & 7 & 8 & 9 & 10 \\
    \midrule
    $x=0$ & (0.2,0.3) & (20,20) & (0.2,0.3) & (20,17) & (17,20) & (20,20) & (0.2,0.4) & (19,12) & (0.2,0.3) & (0.15,0.2) \\
    $x=1$ & (20,20) & (0.2,0.4) & (19,12) & (0.2,0.3) & (0.15,0.2) & (0.2,0.3) & (20,20) & (0.2,0.3) & (20,17) & (17,20) \\
    $x=2$ & (1,10) & (1,10) & (5,10) & (5,2) & (3.1,4) & (19,12) & (0.2,0.3) & (0.15,0.2) & (0.2,0.3) & (20,20) \\
    \bottomrule
  \end{tabular}}
\end{table}

\xhdr{Classifier's Confidence Scores} 
As discussed earlier, the classifier's confidence scores $f_{r}(X_i)$ for each $i \in \Ical, r \in \Rcal$ are generated without access to the subset of patient features $D_i$, and are instead computed using only the observable features $X_i$ and the time slot $r$.
Each score is estimated as the expected value of a Beta distribution parameterized by $X_i$ and time slot $r$, as follows.
\begin{equation}
 f_r(X_i) = \EE_{V \sim \mathrm{Beta}{(\alpha_{X_i,r}, \beta_{X_i,r})}}[V] = \frac{\alpha_{X_i,r}}{\alpha_{X_i,r} + \beta_{X_i,r}}.
\end{equation}


\section{Handling Partial Assignments in the Human Subject Study}
\label{app:exp-nonassign}

To recap, when computing the expected utility of the combined matching outcome, the algorithm first assigned~$n-b$ patients by solving a maximum weight imperfect matching using (noisy) confidence scores, and the remaining $b$ assignments were deferred to a human. The utilities from both components were then combined to obtain the expected utility of the full matching. If a participant was unable to complete all assignments within the $2$-minute time limit, their partial assignment was saved as is. For the experiments in Section~\ref{sec:experiments}, to compute the expected utility of a partial human assignment, we randomly assigned the unmatched patients to the remaining available time slots. While this random assignment is reasonable, to further investigate the impact of different strategies for handling unassigned patients, we explore alternatives to random assignment. Specifically, we examine how the overall utility of the matching changes under these alternative assignment strategies. Additionally, we analyze the effect of filtering out participants who left some instances incomplete, to better understand how partial human assignments influence the expected utility of the combined matching.

\begin{figure}[t]
\centering
\begin{tabular}{c c}
  \textbf{Unassigned} & \textbf{Random assignment} \\
  {\includegraphics[width=.45\textwidth]{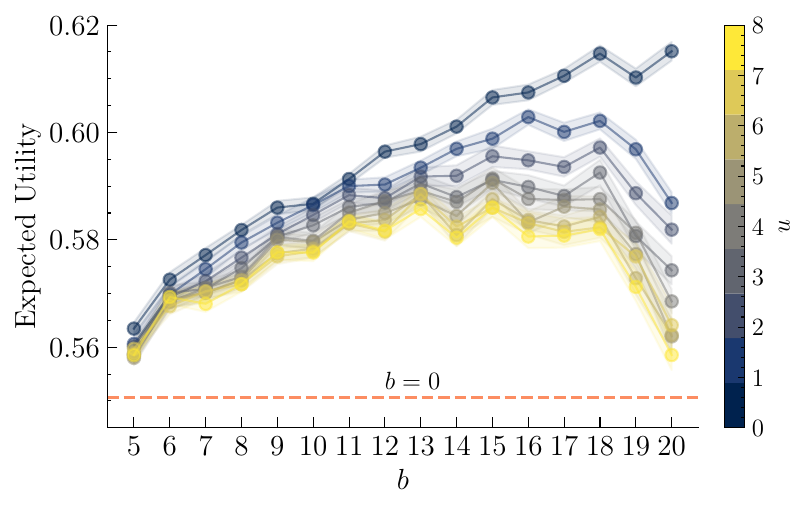}} &
  {\includegraphics[width=.45\textwidth]{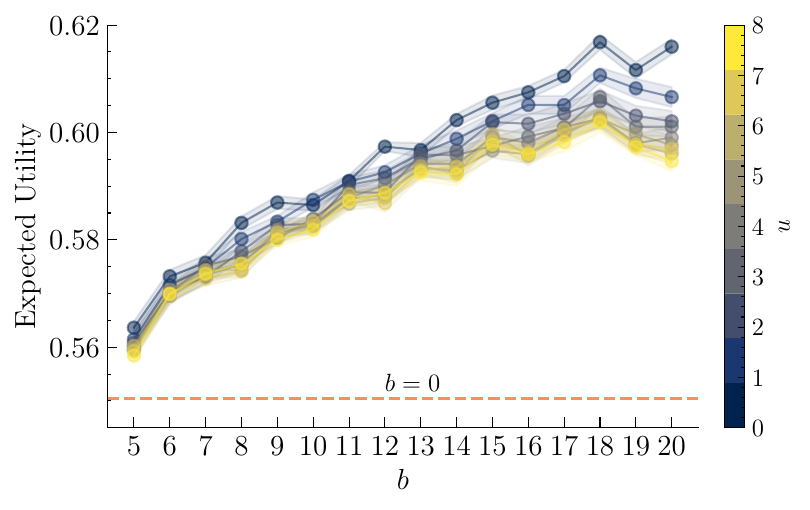}}
\end{tabular}
  \caption{Empirical expected utility per arm achieved by \texttt{UCB1} across $50$ independent realizations, each with a time horizon of $T=2000$. Shaded regions indicate the $95\%$ confidence interval. We progressively remove human participants---along with all their matching assignments---who left more than one patient unassigned in at least $u \in \{0,\dots,8\}$ of the matching assignment tasks, and report the empirical (average) expected utility for each $u$.
  In the left plot, the expected utility is computed by considering the individual success scores of the partial matchings completed by human participants, leaving the unassigned patients excluded while computing the expected utility.
  In the right plot, these unassigned patients are randomly assigned to available time slots, and their corresponding individual success scores are included when computing the expected utility of the combined matching.}
  \label{fig:bandit-filtering}
\end{figure}

\xhdr{Assessing Empirical Expected Utility under Partial Human Assignments}
To understand how partial human assignments affect the empirical expected utility, we conducted an ablation study by progressively excluding participants---and all of their matching assignments---who left more than one patient unassigned in at least $u$ tasks,%
\footnote{If exactly one patient remains unmatched, there is a single available time slot left. In this case, we assign the patient to the available slot, as the choice is unambiguous.}
varying $u \in \{0, \dots, 8\}$. Here, $u = 0$ corresponds to including only participants who completed all assignments in every matching task, while $u = 8$ includes all participants regardless of whether they left any assignments incomplete. 
For the unassigned patients in partial assignments, we considered two strategies. In the first, unassigned patients were left as-is and excluded while computing the expected utility. In the second, they were randomly assigned to available time slots, and the corresponding individual success scores were included to compute the expected utility. In Figure~\ref{fig:bandit-filtering}, we report the empirical expected utility of \texttt{UCB1} for each $b \in \{5,\dots,20\}$---the number of matching decisions deferred to human participants---averaged over $50$ independent realizations. The left plot shows results using partial matchings as-is, while the right plot uses random assignments for unassigned patients. In the second strategy, random assignments were independently resampled for each realization when computing the expected utility. The expected utility of the algorithm-only policy ($b = 0$) is shown using an orange dotted line. However, the expected utility of the human-only policy varies for each $u$, depending on the number of participants retained after filtering. It corresponds to the utility at $b = 20$ for each $u$, \ie, the last point in each curve.

As the value of $u$ decreases, we retain only those participants who consistently completed all assignments. Under these stricter filtering conditions, our framework---splitting the task between humans and the algorithm---achieves higher expected utility across all values of $b$. Notably, when $u = 0$, \ie, considering only the most reliable participants who completed all assignment tasks, the expected utility is close to optimal, even when the matching task with $b=20$ is assigned to these human participants. This reinforces our claim: when human decision-makers are efficient at solving larger matching tasks, deferring more matching decisions to them is more beneficial than relying on the algorithm.
On the other hand, as $u$ increases being less restrictive and we include matchings from under-performing participants, the expected utility consistently decreases---both under partial matchings (with unassigned patients excluded) and under random assignment. This trend is particularly pronounced for higher values of $b$, where a smaller portion of the matching decisions are delegated to human participants.

\begin{figure}[t]
\centering
\begin{tabular}{c c}
 \textbf{Unassigned} & \textbf{Random assignment} \\
 \includegraphics[width=.45\textwidth]{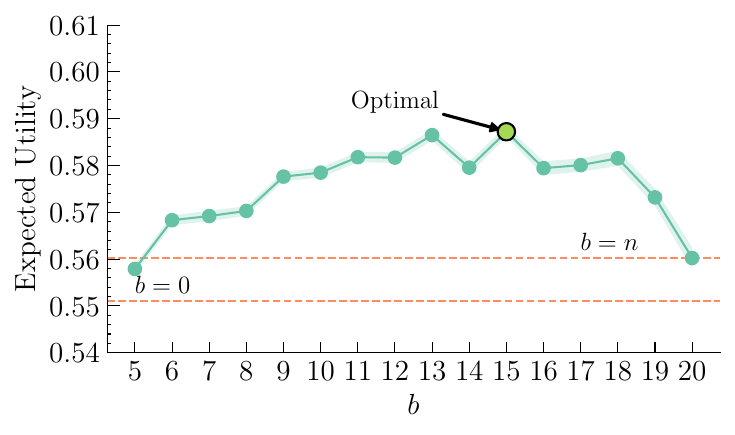} &
 {\includegraphics[width=.45\textwidth] {figs/bandit-results/arm-means/arms-means-h2000-s100-all_nonfiltered_random.pdf}}
\end{tabular}
  \caption{Empirical expected utility per arm achieved by \texttt{UCB1} across $100$ independent realizations, each with a time horizon of $T = 2000$. In the left plot, unassigned patients are left as is and excluded while computing the expected utility. In the right plot, unassigned patients from human partial matchings are randomly assigned to available time slots, and the expected utility of the resulting combined matching is computed. Orange dashed lines indicate the expected utility of the algorithm-only ($b = 0$) and human-only ($b = n$) assignments. Shaded bands represent $95\%$ confidence intervals.}
  \label{fig:bandit-all-unassign}
\end{figure}

\begin{figure}[t]
\centering
\begin{tabular}{c c c}
    \textbf{top-tier} & \textbf{mid-tier} & \textbf{bottom-tier} \\
    {\includegraphics[width=.32\textwidth]
    {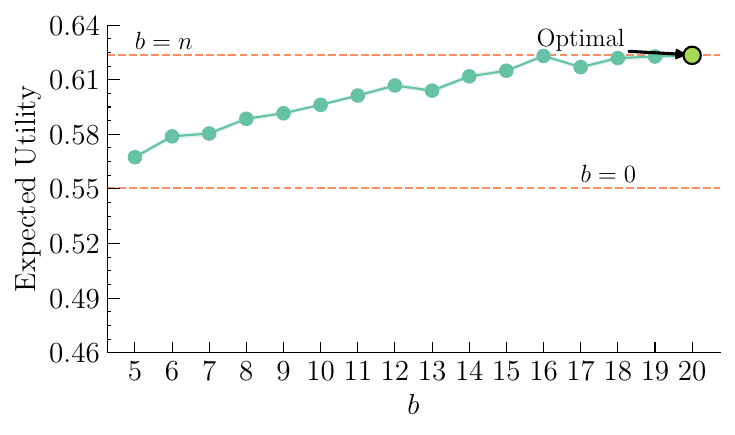}} &
    {\includegraphics[width=.3\textwidth,trim=22 0 0 0,clip]
    {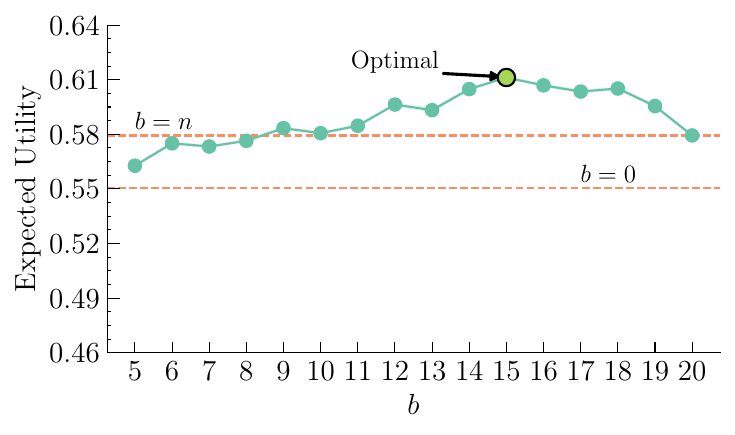}} &
    {\includegraphics[width=.3\textwidth,trim=22 0 0 0,clip]
    {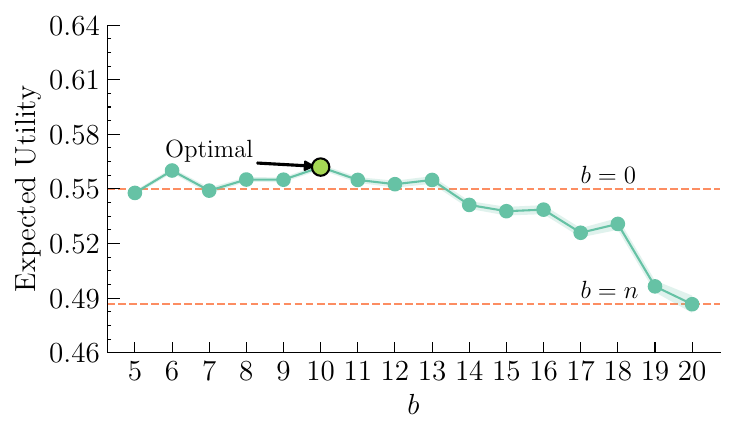}}
\end{tabular}
\caption{Empirical expected utility per arm achieved by \texttt{UCB1} across $100$ different realizations, each of them with an horizon $T=1000$, for a different subset of humans divided by their performance, leaving unassigned
patients as is and excluded from the computation of the expected utility. Shaded bands show 95\% confidence intervals.}
\label{fig:bandit-userGroups-noassign}
\end{figure}

\xhdr{A Fine-Grained Analysis of Partial Human Assignments}
We further examine the case, where all participants---and all of their matching assignments---are included, regardless of whether any patients were left unassigned. Figure~\ref{fig:bandit-all-unassign} compares the expected utility of \texttt{UCB1}, with the left plot showing results where unassigned patients are left as-is, and the right plot showing results where they are randomly assigned to available time slots. 
An immediate observation is that as $b$ increases, deferring more assignments, participants tend to leave more patients unassigned. This reinforces our hypothesis that, when faced with larger matching tasks under limited time, human cognitive limits hinder their ability to complete all assignments.
Further, we observed that the gap between the expected utility of algorithm-only ($b=0$) and human-only ($b=n$) policies is smaller when unassigned patients are left as-is, compared to when they are assigned randomly. This suggests that, as expected, even randomly assigning patients improves the average expected utility, making it a preferable strategy compared to leaving patients unassigned entirely.


We further conducted a more fine-grained analysis by stratifying human participants into three groups---top-, mid-, and bottom-tier---based on the quality of the matchings they produced in the human subject study.%
\footnote{We computed the distribution of expected utility achieved by each participant, divided the full distribution into deciles, and grouped participants accordingly: the bottom $40\%$ were classified as bottom-tier, the middle $20\%$ as mid-tier, and the top $40\%$ as top-tier decision-makers. This resulted in $320$, $160$, and $320$ participants in each group, respectively.}
In Figure~\ref{fig:bandit-userGroups-noassign}, we present the empirical expected utility achieved by \texttt{UCB1} for different values of $b$, by leaving the unassigned patients as-is while computing the expected utility.
As expected, the gap between algorithm-only ($b = 0$) and human-only ($b = n$) assignments is largest for top-tier participants, who outperform the algorithm with a significant margin. This gap narrows for mid-tier participants, but reverses for bottom-tier participants, where human performance is substantially worse than the algorithm-only assignment---as clearly illustrated by the dotted orange lines in the plots.
These findings contrast with the results in Figure~\ref{fig:bandit-usergroups} of the main paper, where---especially among bottom-tier participants who performed poorly---randomly assigning the unassigned patients improved the overall expected utility, underscoring the impact of how unassigned patients are handled.

We further observed that when human participants perform well, deferring more assignments to them is advantageous. This is most evident in top-tier, where the highest expected utility is achieved at $b = 20$, indicating that they can successfully solve the entire matching task --- i.e. all the matchings decisions within a matching task--- without any algorithmic assistance. For mid-tier participants, the peak utility was observed at $b = 15$, while for bottom-tier participants, it drops to $b = 10$. These results reinforce our broader recommendation: as human performance decreases, it becomes less effective to defer a larger share of the task to human decision-makers.

\end{document}